\documentclass[conference]{IEEEtran}
\IEEEoverridecommandlockouts
\usepackage{cite}
\usepackage{amsmath,amssymb,amsfonts}
\usepackage{algorithmic}
\usepackage[caption=false]{subfig}
\usepackage{placeins}
\usepackage{graphicx}
\graphicspath{ {figures/} }
\usepackage{textcomp}
\usepackage{xcolor}
\usepackage[bookmarks=false]{hyperref}
\usepackage{orcidlink}
\def\BibTeX{{\rm B\kern-.05em{\sc i\kern-.025em b}\kern-.08em
    T\kern-.1667em\lower.7ex\hbox{E}\kern-.125emX}}
    
    \makeatletter
\def\ps@IEEEtitlepagestyle{
  \def\@oddhead{\mycopyrightnotice}
  \def\@evenhead{}
}
\def\mycopyrightnotice{
  {\footnotesize
  \begin{minipage}{\textwidth}
  \copyright\  2022 IEEE.  Personal use of this material is permitted.  Permission from IEEE must be obtained for all other uses, in any current or future media, including reprinting/republishing this material for advertising or promotional purposes, creating new collective works, for resale or redistribution to servers or lists, or reuse of any copyrighted component of this work in other works.\\
  DOI: 10.1109/ICMLA55696.2022.00012\hfill Published version: https://ieeexplore.ieee.org/document/10068967
  \end{minipage}
  }
}

\begin{document}

\title{An Empirical Evaluation of Multivariate Time Series Classification with Input Transformation across Different Dimensions
\thanks{This work has been conducted as part of the Just in
Time Maintenance project funded by the European Fund for
Regional Development.}
}

\author{\IEEEauthorblockN{Leonardos Pantiskas}
\IEEEauthorblockA{\textit{Vrije Universiteit Amsterdam}\\\orcidlink{0000-0002-4898-5334} 0000-0002-4898-5334}
\and
\IEEEauthorblockN{Kees Verstoep}
\IEEEauthorblockA{\textit{Vrije Universiteit Amsterdam}\\\orcidlink{0000-0001-6402-2928} 0000-0001-6402-2928}
\and
\IEEEauthorblockN{Mark Hoogendoorn}
\IEEEauthorblockA{\textit{Vrije Universiteit Amsterdam}\\
\orcidlink{0000-0003-3356-3574} 0000-0003-3356-3574}
\and
\IEEEauthorblockN{Henri Bal}
\IEEEauthorblockA{\textit{Vrije Universiteit Amsterdam}\\
\orcidlink{0000-0001-9827-4461} 0000-0001-9827-4461}
}

\maketitle

\begin{abstract}
In current research, machine and deep learning solutions for the classification of temporal data are shifting from single-channel datasets (univariate) to problems with multiple channels of information (multivariate). The majority of these works are focused on the method novelty and architecture, and the format of the input data is often treated implicitly. Particularly, multivariate datasets are often treated as a stack of univariate time series in terms of input preprocessing, with scaling methods applied across each channel separately. In this evaluation, we aim to demonstrate that the additional channel dimension is far from trivial and different approaches to scaling can lead to significantly different results in the accuracy of a solution. To that end, we test seven different data transformation methods on four different temporal dimensions and study their effect on the classification accuracy of five recent methods. We show that, for the large majority of tested datasets, the best transformation-dimension configuration leads to an increase in the accuracy compared to the result of each model with the same hyperparameters and no scaling, ranging from 0.16 to 76.79 percentage points. We also show that if we keep the transformation method constant, there is a statistically significant difference in accuracy results when applying it across different dimensions, with accuracy differences ranging from 0.23 to 47.79 percentage points. Finally, we explore the relation of the transformation methods and dimensions to the classifiers, and we conclude that there is no prominent general trend, and the optimal configuration is dataset- and classifier-specific.
\end{abstract}


\section{Introduction}
Due to the rising availability of data sources in sectors such as industry, healthcare, and finance, time series classification datasets and problems are increasingly consisting of multiple channels of information \cite{ruizGreatMultivariateTime2021}, representing for instance readings from multiple sensor types. Following this trend, machine and deep learning solutions have shifted to try to more effectively address these multivariate problems \cite{ruizGreatMultivariateTime2021}.

An aspect that does not usually get much focus when describing a novel machine or deep learning solution is the preprocessing of the input data. With the terms preprocessing, scaling, and transformation we refer to the methods that modify a set of values to deal with issues such as outliers or to shift them to a predefined range, e.g. standardization or min-max scaling. Especially in the deep learning landscape, it is now a standard approach at the implementation level to implicitly transform all values with some method such as normalization, in order to bring them to the same numerical scale prior to propagating them through the network.
In the time series classification field, there has already been extended research for the univariate datasets \cite{bagnallGreatTimeSeries2017}, so it is a natural direction to try and transfer the concepts to the multivariate cases. This can regard the models themselves, e.g. by applying a model to each univariate channel of a multivariate problem and then aggregating the classification decisions using some ensembling or voting method \cite{ruizGreatMultivariateTime2021}. Similarly to this approach, the preprocessing of the input also follows the same pattern. For example, if in the case of univariate classification, where the dataset dimensions are (\textit{samples},\textit{timesteps}), all observations are standardized, it seems natural to apply the same transformation to each channel of the multivariate dataset which has dimensions (\textit{samples},\textit{channels},\textit{timesteps}), treating it as a separate univariate entity.
However, as has been noted by Ruiz et al. in \cite{ruizGreatMultivariateTime2021}, the transformation of the input data in the case of multivariate datasets is not a trivial problem. The additional data dimension presents several options even in the fundamental handling of this prior input scaling.

In this work, we want to empirically explore the opportunities that can arise from scaling the temporal data across different dimensions and the effect this can have on classification accuracy. In order to achieve this, we experiment with five recent multivariate time series classification models, which are a mix of deep learning and other methods and are representative solutions with results equal to or sufficiently close to the state of the art. We choose seven different transformation methods and apply them to four distinct slices of the temporal data. Our contributions are:
\begin{itemize}
\item We show that in the large majority of datasets tested, the best combination of transformation method and dimension it is applied to leads to better accuracy than that of the models with the same functional hyperparameters and no input scaling, ranging from 0.16 to 76.79 percentage points.
\item We show that in the majority of the best configurations, there is a statistically significant difference among the accuracy results of the models when the same transformation method is applied to different temporal dimensions, ranging from 0.23 to 47.79 percentage points.
\item We explore the relation of the best transformation methods and dimensions to the models and we find that there are no distinct general trends and that the best configuration depends on the dataset and classifier used.
\end{itemize}

\section{Related Work}

In fields such as computer vision, data-centric approaches that aim to increase model accuracy have been widely utilized, with data augmentation methods such as cropping and rotation applied to images to increase the amount of the training data, tackle class imbalances, and make the model more robust to perturbations of the input\cite{shorten2019imagedataaugm}. In contrast to that, data augmentation techniques in the time series domain have been less extensively employed. Although the three-dimensional nature of the multivariate time series problem may initially resemble image data, the temporal dependencies and dynamics of channels, as noted in \cite{wen2021survey}, lead to a qualitative difference between the challenges.

In recent surveys of time series data augmentation \cite{iwana2021empiricaltimeaugm,wen2021survey} the methods presented range from basic ones inspired by the computer vision field, such as flipping and slicing samples, to more advanced ones utilizing deep generative models. In a recent work specifically focused on multivariate time series classification\cite{yang2022robust}, the authors showed that basic time series augmentation methods can be beneficial to the task, counter-acting the overfitting of models, especially on smaller datasets.

Our exploration of the transformation of data across different dimensions, although not strictly a data augmentation method, can be considered a data-centric approach, in the sense that we are trying to achieve better classification accuracy only by modifying the input data in a specific manner. It is orthogonal to the above data augmentation methods and is conceptually placed at an earlier stage. It stems from a consideration of the inherent nature of the time series datasets and what the intrinsic relationship of each dimension slice is to the real-world problem.
\section{Methodology}
\subsection{Models}
The landscape of multivariate time series classification models is continuously evolving, with ever-more complex and accurate models and methods \cite{ruizGreatMultivariateTime2021}. We can, however, distinguish two broad categories which encompass multiple recent methods: machine learning models based on extracted features and deep learning models. In the first case, several features are extracted from either the input values or a transformation of them, and those features are then used with a linear classifier for the final classification. In the second category, the input values are propagated to a deep learning architecture, usually after normalization to bring them to a similar scale. The architecture then internally performs the end-to-end transformation and classification of the input. We select three recent methods belonging to the first category and two belonging to the second. A short description of the methods follows, starting with the feature extraction ones:

\textbf{ROCKET}\cite{dempster2020rocket} is a method based on random convolutional kernels which not only achieves the best results in terms of accuracy according to a recent evaluation \cite{ruizGreatMultivariateTime2021} but is also the fastest approach in terms of training time. Two features are extracted from the output of the convolution of the input with each random kernel.

\textbf{WEASEL+MUSE}\cite{schaferMultivariateTimeSeries2018} extracts features from windows of the input, utilizing a truncated Fourier transform and bag-of-patterns approach. It also applies statistical filtering of these features using a $\chi^{2}$ test.

\textbf{LightWaveS}\cite{pantiskaslightwaves2022} utilizes lightweight wavelet scattering with arbitrary wavelets. Four statistical features are extracted from each of the scattering coefficients and are filtered with a hierarchical feature selection approach.

Our selected models in the deep learning category are:

\textbf{ResNet}\cite{wang2017resnet}, which has been proposed as a strong deep learning baseline for the time series classification task, with its architecture consisting of convolutional layers, residual connections, and global average pooling layers.

\textbf{InceptionTime}\cite{ismail2020inceptiontime}, which utilizes a more complex architecture of convolutional blocks and bottleneck layers, in the form of Inception modules \cite{szegedy2015inceptionmodule}, with residual connections. 

The models were selected on the basis of them being recent time series classification methods, which were designed to handle multivariate data and are not just ensembles of univariate methods. Moreover, the selected models include the best classifiers for 20 out of the 26 equal-length UEA problems according to the reported accuracy metrics in \cite{ruizGreatMultivariateTime2021}, so they are a very representative sample of the current state of the art. Another factor that was taken into account was the computational cost of each method. Since we have to perform multiple resamplings of multiple transformation methods across four data slices, we have to limit the model selection in order for the experiments to finish within a reasonable amount of time. Thus, although solutions such as HIVE-COTE \cite{lines2018time} and CIF \cite{middlehurst2020canonical} may rank higher in accuracy for some problems, their very long training time makes it impractical to fairly include them in our evaluation.

\subsection{Transformation methods}
Data scaling is of course a standard method during the exploratory data analysis phase of a problem. However, as we mentioned above, it is often overlooked when presenting novel time series classification approaches, especially in the recent deep learning environment, where the model architecture is expected to reach the correct weight values regardless of the input format. In our experiments, we test seven different well-known transformation methods\cite{scalingmethods}, ranging from simple linear to more complex non-linear functions. We present those below, with a short description for the sake of completeness:

\textbf{Normalization} The values are transformed so that their L2 norm is 1.

\textbf{Standardization} The values are transformed so that they have zero mean and unit variance.

\textbf{MinMax} The values are scaled to the [0,1] range.

\textbf{MaxAbs} The values are scaled based on their maximum absolute value, but their sign is retained. On positive data, this method is equivalent to MinMax.

\textbf{Robust} The values are scaled based on their median and interquartile range, which are robust against outliers.

\textbf{Power Transformation} The values are non-linearly transformed with a power transformation (Yeo-Johnson method in our experiments) in order to approach a Gaussian distribution, minimizing skewness and stabilizing variance.

\textbf{Quantile Transformation} The values are non-linearly transformed so that their probability density function is mapped to a uniform distribution with a [0,1] range.

\subsection{Dimensions}

The 3-dimensional nature of multivariate time series problems, namely samples, channels, and timesteps, presents a multitude of options for selecting data slices across which the appropriate transformation method can be applied. As we said, as a result of the mapping of concepts from the univariate time series research, it may seem natural to transform all values of each channel as a separate set. In this work, we claim that this choice is not standard or trivial and that different configurations may lead to considerably different results. For instance, it is generally accepted that a large part of the added value in multivariate datasets comes from the interplay and associations among different channels. Thus, by selecting data slices that include values across different channels, we introduce such associations even before the processing of the input by the models, which may be able to help them perform better in some datasets. We denote the original dataset as $D$ with $N$ samples, $C$ channels, and $T$ timesteps. Below we present the four distinct data slices that we selected for experimentation, along with an intuitive explanation:

\textbf{Channels} This is the configuration more closely related to the univariate paradigm, as all values of each channel across all samples are considered a separate set $S_{i} = \{ D_{*,i,*}\}, 1\leq i\leq C$.

\textbf{Timesteps} In this configuration, the values of each timestep across all samples and all channels are considered a separate set $S_{i} = \{ D_{*,*,i}\}, 1\leq i\leq T$.

\textbf{Both} This configuration is a combination of the above, where for each channel, the values of each timestep across all samples are considered a separate set $S_{ij}= \{D_{*,i,j} \}, 1\leq i\leq C , 1\leq j \leq T$.

\textbf{All} In this configuration, all values of the dataset are taken as a single set $S = \{D_{*,*,*}\}$.

This non-exhaustive selection of dataset slices is based on the rationale of capturing the intuitive, real-world meaning of each dimension. For example, if the different channels represent sensors with significantly different value ranges, it would make sense to transform them separately. On the other hand, if all sensors are of the same type, but for instance, their readings come from fixed points of a process, then each timestep is potentially a more important dimension to consider for classification. By combining these concepts, we end up with the four slices mentioned above. We see that two of our selected slices do not include information sharing across time series (\textbf{channels} and \textbf{both}) while the other two do. 

\section{Experiments}
\subsection{Datasets}
We experiment on the 26 of the 30 datasets of the UEA collection \cite{bagnallUEAMultivariateTime2018} that have equal-length samples and can thus be handled easily by all models. Moreover, this is the same subset for which there are detailed metrics in \cite{ruizGreatMultivariateTime2021}, so we have a robust point of reference. For the WEASEL+MUSE method, we also exclude the datasets \textit{DuckDuckGeese}, \textit{EigenWorms}, \textit{FaceDetection}, \textit{MotorImagery}, \textit{PEMS-SF}, and \textit{PhonemeSpectra}, due to its inability to successfully complete the training on these, as also noted in \cite{ruizGreatMultivariateTime2021}.

\subsection{Experimental setup}
All experiments were run on the DAS-6 infrastructure \cite{balDAS2016}, on nodes with 24-core AMD EPYC-2 (Rome) 7402P CPUs, NVIDIA A6000 GPUs, and 128 GB of RAM. 
We implement ROCKET and MUSE using \textit{sktime}\cite{sktime} and InceptionTime and ResNet using its deep learning extension, \textit{sktime-dl}. For LightWaveS, we use its provided code.

Regarding the method parameters, we tried to follow as closely as possible the ones reported in \cite{ruizGreatMultivariateTime2021} and used the default settings for ROCKET and LightWaveS. We present those parameters in detail in Table~\ref{table:model_params}. As baseline, we use the models on the unmodified UEA datasets and in addition, we disable all data preprocessing in the methods that allow this. 

We used scikit-learn \cite{scikit-learn} to implement all scaling methods. We repeat each experiment 20 times with different starting seeds and get the mean accuracy. For each model and dataset, we sort the different configuration results by descending mean accuracy and then ascending standard deviation among resamples. In this way, we find the scaling method - dimension combination which yields the highest mean accuracy and most stable results. We present the mean-accuracy difference from the baselines only when it is statistically significant.

In order to distinguish the value of dimension selection from that of the transformation method, we present another set of results: For each of the datasets and models, we keep the transformation method of the best-performing configuration fixed and we apply it to all four data slices. We then do pairwise testing to determine the statistical difference between all possible pairs of the four accuracy results. 

We do not make any assumptions about the result distributions, so in both cases, we use the Wilcoxon signed-rank test\cite{wilcoxon1992individual} with p-value of 0.05 and Holm’s alpha correction when needed\cite{holm1979simple}. The code for the experiments as well as the detailed metrics are made available at \url{https://github.com/lpphd/mtsscaling} to facilitate reproducibility of the results.

\begin{table}[ht]
\caption{Method parameters}\label{table:model_params}
\begin{center}
\begin{tabular}{|p{0.2\linewidth}|p{0.7\linewidth}|}
\hline
Method &  Parameters\\
\hline
ROCKET &  Ridge regression classifier, 10000 kernels \\\hline
WEASEL+ MUSE &  Default sktime parameters (anova=True, bigrams=True, window\_inc=2, p\_threshold=0.05, use\_first\_order\_differences=True) \\\hline
LightWaveS &  Ridge regression classifier, 500 features \\\hline
ResNet &  Epochs: 1500, Batch size: 16,\\& Learning rate: 1e-3 and  halved after no improvement for 50 epochs\\
& Three residual blocks each with three conv layers with kernel sizes [8, 5, 3]\\&
Filters per conv layer for each block [64, 128, 128] \\
& Training ends after no improvement for 150 epochs
\\& Weights with lowest training loss are used for testing\\\hline
InceptionTime &  Epochs: 1500, Batch size: 16,\\& Learning rate: 1e-3 and  halved after no improvement for 50 epochs\\
& Two residual blocks each with three Inception modules with kernel sizes per module [10, 20, 40]\\&
Plus bottleneck filters for all conv layers 32 \\
& Training ends after no improvement for 150 epochs
\\& Weights with lowest training loss are used for testing\\
\hline
\end{tabular}
\end{center}
\end{table}

\section{Results}

 We present the best-achieved accuracy, as well as the difference from the baseline result for each model in Table~\ref{table:new_sota}.

\begin{table}[ht]
\caption{Accuracy under best transformation-dimension configuration and significant differences from baseline accuracy}\label{table:new_sota}
\begin{center}
\begin{tabular}{|c|c|c|c|c|c|}
\hline
 & ROCKET   & MUSE     & LightWaveS & ResNet   & IT \\ \hline
AWR     & 99.8     & 99.3     & 99.7       & 98.2     & 98.7          \\ 
        & (+0.48)  & (-)  & (-)    & (+0.35)  & (+0.3)       \\ \hline
AF      & 46.7     & 40.7     & 46.7       & 38       & 36            \\ 
        & (+26.67)  & (+15.0) & (+13.33)   & (+8.0)   & (+15.67)      \\ \hline
BM      & 100      & 100      & 100        & 100      & 100           \\ 
        & (-)    & (-)    & (-)      & (-)    & (-)         \\ \hline
CR      & 100      & 100      & 97.2       & 99.9     & 99.4          \\ 
        & (-)    & (+0.62)  & (+4.17)    & (+1.18)  & (+0.83)       \\ \hline
DDG     & 69.1     & N/A      & 52         & 69       & 66.8          \\ 
        & (+7.4)   &          & (+8.0)     & (+7.7)   & (+7.7)        \\ \hline
ER      & 98.6     & 97       & 96.7       & 93.1     & 91.2          \\ 
        & (-)  & (+1.52)  & (-0.37)    & (+6.48)  & (+2.74)       \\ \hline
EW      & 96.5     & N/A      & 96.2       & 93.9     & 94.9          \\ 
        & (+6.07)  &          & (-)    & (+76.79) & (+10.53)      \\ \hline
EP      & 100      & 100      & 97.8       & 99.2     & 97.9          \\ 
        & (-)    & (-)    & (+1.45)    & (+0.43)  & (+1.01)       \\ \hline
EC      & 44.7     & 37.8     & 63.5       & 30.4     & 29.6          \\ 
        & (-)  & (-)   & (-0.38)    & (+5.91)  & (+2.45)       \\ \hline
FD      & 66.1     & N/A      & 64.5       & 65.7     & 66.7          \\ 
        & (+2.3)  &          & (+3.2)    & (+7.49)  & (+1.83)        \\ \hline
FM      & 59.8     & 57       & 59         & 56       & 59.2          \\ 
        & (+5.7)   & (+3.8)   & (+4.0)     & (+2.2)  & (-)        \\ \hline
HMD     & 50.1     & 34.6     & 43.2       & 36.2     & 44.7          \\ 
        & (-)  & (+3.78)  & (+9.46)    & (-)  & (+7.57)       \\ \hline
HW      & 59       & 38.2     & 38.2       & 61       & 58.2          \\ 
        & (+0.46)  & (+9.76)  & (+1.18)    & (+1.85)  & (-)       \\ \hline
HB      & 78.9     & 79.5     & 80.5       & 77.5     & 76.7          \\ 
        & (+1.9)  & (+6.2)  & (+7.32)    & (+20.78) & (+7.68)        \\ \hline
LSST    & 67.8     & 64.7     & 47.8       & 61.6     & 64.5          \\ 
        & (+3.49)  & (+4.04)  & (+13.76)   & (+10.0)  & (+13.53)       \\ \hline
LIB     & 96.2     & 91.9     & 90         & 96.1     & 90.1          \\ 
        & (+2.33)  & (-)  & (+5.0)     & (+0.5)  & (+1.06)        \\ \hline
MI      & 59.6     & N/A      & 64         & 54.6     & 52.4          \\ 
        & (+5.45)   &          & (+13.0)    & (+2.1)  & (+2.05)        \\ \hline
NATO    & 95.1     & 93.7     & 76.7       & 97.4     & 95.7          \\ 
        & (+1.14)  & (+1.53)  & (+13.33)   & (+1.06)  & (-)       \\ \hline
PEMS    & 88.3     & N/A      & 92.5       & 91.4     & 87.7          \\ 
        & (+15.32) &          & (+11.56)   & (+13.5) & (+12.31)      \\ \hline
PD      & 98.3     & 94.8     & 95.4       & 98.6     & 98.9          \\ 
        & (+0.16)  & (+0.22)  & (-0.05)    & (-)  & (-)       \\ 
        \hline
PS      & 31.6     & N/A      & 22.7       & 31.6     & 30.7          \\ 
        & (+0.65)  &          & (+7.58)    & (-)  & (+0.63)       \\ \hline
RS      & 93       & 89.9     & 88.8       & 93       & 92.9          \\ 
        & (+3.16)   & (+2.53)  & (+5.26)    & (+1.97)  & (+3.88)       \\ \hline
SRS1    & 93       & 77.5     & 88.4       & 78.6     & 86.3          \\ 
        & (+2.2)  & (-)  & (+9.22)    & (+1.57)  & (+1.43)       \\ 
        \hline
SRS2    & 57.5     & 55.4     & 53.9       & 52.3     & 54            \\ 
        & (+6.64)  & (+2.97)  & (+7.22)    & (+5.19)  & (+4.25)       \\ \hline
SWJ     & 51.3     & 53.3     & 60         & 41       & 70            \\ 
        & (+3.33)   & (+11.0) & (+6.67)    & (+12.33) & (+34.0)       \\ \hline
UW      & 94.3     & 93       & 94.1       & 86.6     & 90            \\ 
        & (+0.61)  & (+1.62)  & (+3.12)    & (+0.98)  & (+0.5)       \\ \hline
\end{tabular}
\end{center}
\end{table}

We can see that there is an increase in accuracy in the large majority of the datasets for all models, ranging from 0.16 to 76.79 percentage points, with the median increase being 3.88 percentage points. Broken down by model, the median increase in accuracy is 2.75 for ROCKET, 3.38 for MUSE, 7.27 for LightWaveS, 3.7 for ResNet, and 2.74 for InceptionTime. It is remarkable that not only is the accuracy increased compared to the baseline experiments, but the best accuracy across all models is higher than the best accuracy presented in \cite{ruizGreatMultivariateTime2021} for 13 out of the 26 datasets, showing that this input preprocessing exploration can result in new state-of-the-art results without modifying the base model at all.

A point that merits explanation is that this approach of no input scaling as baseline differs from the default behavior of models such as ROCKET and LightWaveS, which employ scaling as part of their pipeline, or the usual normalization for deep learning models. The reason we follow it is to get as fair results as possible and create a reference point based only on the mechanics of the models rather than any scaling effect. However, we can confirm that the same trends and conclusions hold true for the default behavior of our selected classifiers by getting their reported accuracy metrics on the same datasets from \cite{ruizGreatMultivariateTime2021},\cite{pantiskaslightwaves2022} and performing mean-accuracy comparison. Again, for the majority of datasets and classifiers, there is an increase in accuracy, ranging from 0.1 to 40.0 with a median of 3.25.

There are also a few negative results, which indicate that the application of no transformation gives better accuracy for specific models and datasets. These results do not affect our conclusions, since we are considering the transformation method and dimension as hyperparameters, and we can include additional configurations in this hyperparameter search to increase the chances of achieving the optimal result.

We also aim to distinguish the value of dimension selection from that of the transformation method. Although in practice the scaling method and dimension would be co-selected based on their interplay and the dataset characteristics, we want to demonstrate that even for more complex transformation methods, the dimension selection can significantly affect the outcome. In Table~\ref{table:dim_diff} we see whether or not there is a statistically significant difference in the accuracy results between any two dimensions under the optimal scaling method, and if so, what the difference is in the mean accuracy between the optimal and worst dimension.

\begin{table}[ht]
\caption{Difference (of significantly different results) in mean accuracy between best and worst dimension for fixed transformation method}\label{table:dim_diff}
\begin{center}
\begin{tabular}{|c|c|c|c|c|c|}
\hline
     & ROCKET & MUSE  & LightWaveS & ResNet & IT    \\ \hline
AWR  & 0.42   & 1     & 1.33       & 0.63   & -     \\ \hline
AF   & 33.33  & 15    & 26.67      & 7.33   & 12.33 \\ \hline
BM   & -      & -     & -          & -      & -     \\ \hline
CR   & 1.39   & 2.78  & 6.94       & 3.82   & 1.53  \\ \hline
DDG  & 8      & N/A   & 32         & 10.6   & 3.1   \\ \hline
ER   & 1.76   & 2.46  & 1.48       & 6.85   & 3.41  \\ \hline
EW   & 47.79  & N/A   & 14.5       & 3.05   & 5.27  \\ \hline
EP   & -      & 0.72  & 1.45       & 0.65   & 4.13  \\ \hline
EC   & 10.32  & 5.21  & 20.91      & 2.49   & -     \\ \hline
FD   & 1.02   & N/A   & -          & 1.77   & -     \\ \hline
FM   & 6.05   & 4.45  & 12         & 5.35   & -     \\ \hline
HMD  & 6.35   & -     & 24.32      & 10     & 10.95 \\ \hline
HW   & 11.74  & 3.2   & 13.29      & 16.99  & 19.25 \\ \hline
HB   & 3.73   & -     & 5.85       & -      & 2.56  \\ \hline
LSST & 0.41   & 13.42 & 0.66       & 3.73   & -     \\ \hline
LIB  & 1.72   & 1.36  & 2.22       & -      & -     \\ \hline
MI   & 7.95   & N/A   & 15         & 2.45   & 2.25  \\ \hline
NATO & 2.39   & 8.14  & 19.44      & 3.78   & 1.11  \\ \hline
PEMS & 5.84   & N/A   & 8.67       & 6.94   & 9.48  \\ \hline
PD   & 0.31   & 0.69  & -          & 0.23   & -     \\ \hline
PS   & 0.56   & N/A   & 7.15       & 1.94   & 1.76  \\ \hline
RS   & 0.63   & 3.06  & 10.53      & 1.51   & 2.47  \\ \hline
SRS1 & 0.84   & -     & 11.95      & -      & 2.83  \\ \hline
SRS2 & 3.31   & 3.31  & 5.56       & 4.67   & 3.14  \\ \hline
SWJ  & 10.33  & 11.33 & 13.33      & 7      & 33    \\ \hline
UW   & 0.42   & 5.5   & 1.88       & -      & -     \\ \hline
\end{tabular}
\end{center}
\end{table}

These results reinforce our conclusions, as we can see that for the majority of datasets and models there is a statistically significant difference between the results of at least two out of the four dimensions and the mean-accuracy differences range from 0.23 to 47.79 percentage points, with the median being 3.82 points. This shows that a significant part of the accuracy increase compared to the baselines stems from the selection of the most suitable dimension for a given dataset and classifier.

\begin{figure}[hbtp]
    \centering
  \subfloat[Dimension scores\label{fig:dimensions_scores}]{%
       \includegraphics[width=1\linewidth]{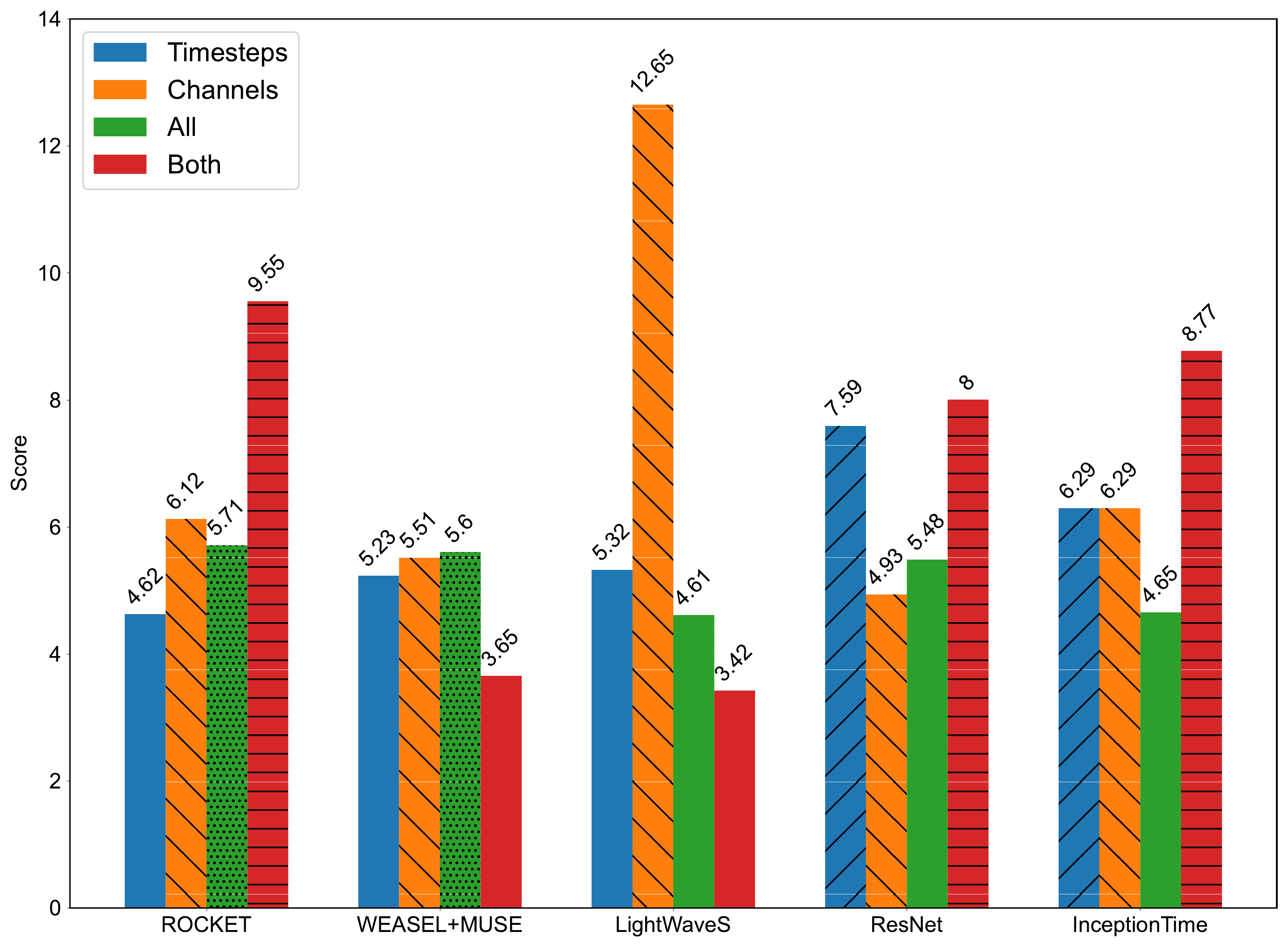}}    
    \\
  \subfloat[Transformation method scores \label{fig:method_scores}]{%
        \includegraphics[width=1\linewidth]{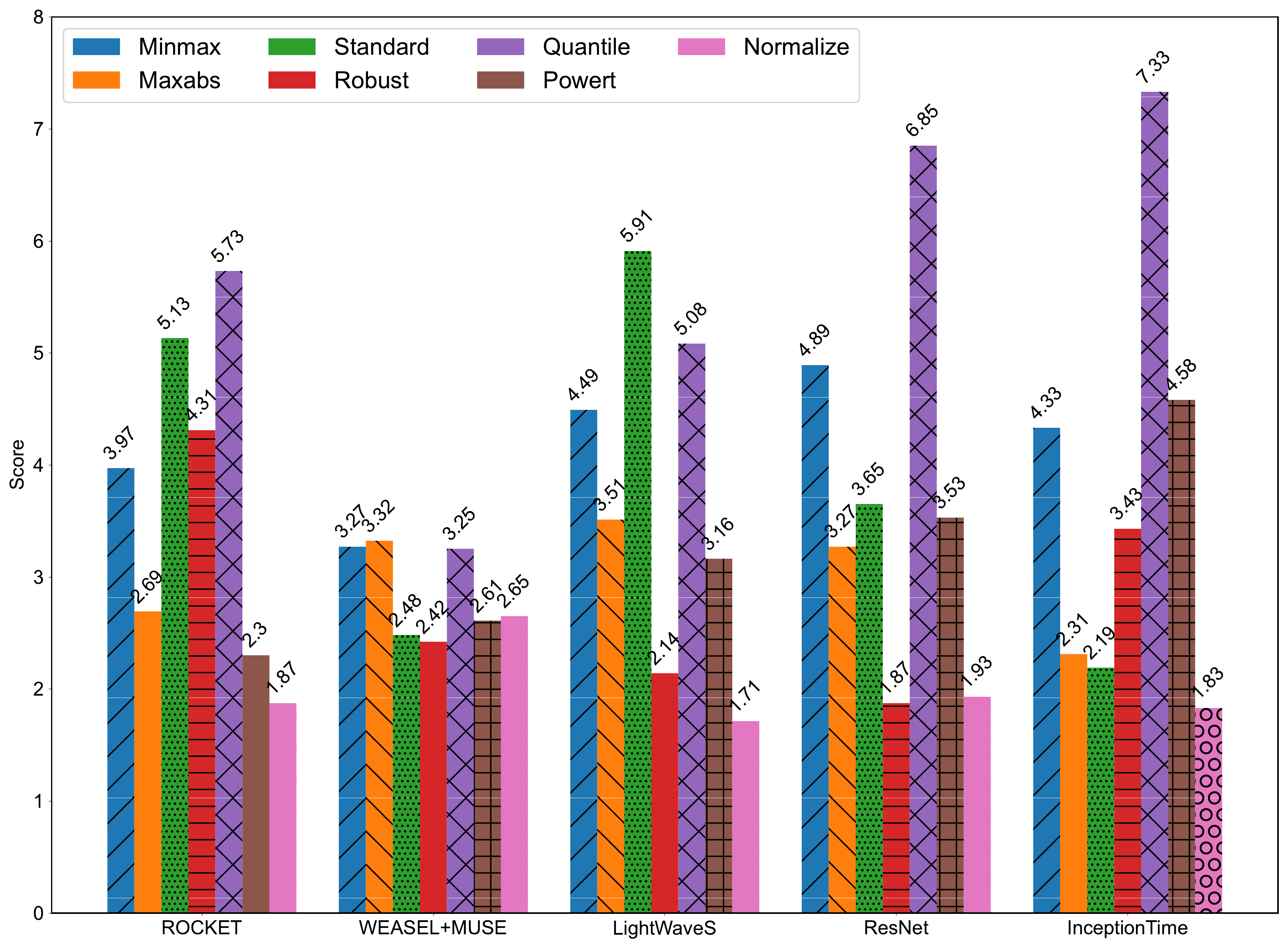}}
\caption{Utility scores of dimensions and transformation methods for each classifier}
             \label{fig:scores}
\vspace{-1em}
\end{figure}

We can also study the configurations that achieve the best performances to discover potential trends in the dimension or transformation method selection. To achieve this, for each classifier and dataset we consider the group of configurations that help achieve either the top accuracy or within 1 percentage point of it. We then calculate a score for each dimension and transformation method that appears in these configurations, which is defined as the number of times it appears divided by the total number of the group members. For example, if the top configurations are [minmax\_both, standard\_both, quantile\_all], the dimension 'Both' would get a score of (1+1)/3 = 2/3, while each of 'MinMax', 'Standard', 'Quantile' methods would get a score of 1/3.
By summing this normalized score across all datasets, we get a "usefulness" profile of the dimensions and transformation methods for each classifier.
We can see these scores in Fig. ~\ref{fig:scores}.

Regarding the dimension scores in Fig. ~\ref{fig:dimensions_scores}, we can see that there is no universal trend and the results are classifier-specific. One result that stands out is the high utility of "Channel" dimension for the LightWaveS method. This seems to be related to the method's mode of operation, which is extracting features from individual input channels, without combining them in any way. ROCKET does combine channels when generating features, and its top dimension is "Both". This dimension is also the top one for the two deep learning models, although for ResNet the "TimeSteps" dimension is also valuable.
On the contrary, in WEASEL+MUSE we observe that "Both" has the lowest value, with a balance across the other dimensions.

Considering the transformation method scores in Fig. ~\ref{fig:method_scores}, we see again that the results vary across classifiers. The two deep learning models show a common behavior in that "Quantile" and "MinMax" are valuable methods in both, which seems to validate the common approach of scaling the input to the [0,1] range. However, they also show deviation in the scores of methods such as "Robust" and "Standard". "Quantile" is also the top method for ROCKET, followed by "Standard", while for LightWaveS this order is reversed. MUSE seems to have a relative balance across transformation methods. A general trend that we can observe is that the "Quantile" method seems to be useful for all classifiers, while the simple normalization method has a low score in all cases.

These figures show that there is no clear winner either in dimension or transformation methods, and the best configuration depends on the classifier and the dataset under consideration. The conclusion we can draw from this is that the inclusion of data dimension and transformation method in the hyperparameter search of a model is the most certain method of discovering the configuration that gives the optimal result in terms of accuracy.

\section{Discussions and Conclusion}
In summary, in this paper, we empirically explore the input preprocessing possibilities presented by the format of multivariate time series datasets, namely \textit{(samples, channels, timesteps)} and their effect on classification accuracy. We test seven data transformation methods across four distinct data slices and their effect on the accuracy of five recent machine and deep learning methods. We show that the optimal configuration of data slice and transformation method leads to an increase in the classification accuracy in almost all cases, in comparison with the baselines without any preprocessing. We also show that the correct dimension selection can lead to a large accuracy increase compared to a sub-optimal selection.

The above empirical results affect topics on a broad spectrum of time series analysis and classification, from the evaluation of novel methods to computational cost savings. In research, these results indicate that data-centric approaches are a fruitful research direction and can have significant benefits in terms of classification accuracy, on par or even better than new methods, without incurring the additional model complexity, especially in the landscape of deep learning. In this case, the novel approaches should be evaluated based on additional aspects, such as interpretability or deployment suitability. Similarly, in the more applied industry sector, it points practitioners to the possibility of increasing accuracy for a specific use case through data-centric means, obviating the need to switch to more computationally expensive or communication-intensive models, especially in the edge intelligence applications.
A natural research direction stemming from our work is to formalize the discovery of the most suitable transformation method and dimension. Although for the faster methods such as ROCKET and LightWaveS it is easy to quickly search for the best configuration, it is quite impractical for the slower methods such as MUSE. Thus, a desirable approach would indicate the optimal dimension for each dataset, possibly based on the statistical properties of each data slice, and also explain this choice.
In terms of more applied directions, it would be interesting to experiment with additional models which may have more markedly different approaches than the ones presented, such as shapelet-based ones\cite{ruizGreatMultivariateTime2021}. Finally, additional data slices could be explored, such as grouping channels depending on the underlying problem and data source type, e.g., sensors of similar type in an IoT problem.

\bibliographystyle{IEEEtran}
\bibliography{IEEEabrv,references}

\end{document}